\documentclass[runningheads]{llncs}

\usepackage[hidelinks]{hyperref}
\usepackage{graphicx}
\usepackage{amsmath}
\usepackage{booktabs}
\usepackage{amsfonts,amssymb}
\usepackage{multirow}
\usepackage{multicol}
\usepackage{subfigure}
\usepackage{color}
\usepackage{colortbl,algorithm2e,enumitem}%,algorithmic}
\usepackage{todonotes}
\usepackage{booktabs}
\usepackage{textgreek}
\usepackage{comment}
\usepackage{wrapfig,lipsum,booktabs}

\newcommand{\m}{{\sf {MWSS}}}
\newcommand{\gn}{{\sf {LWN}}}

\newcolumntype{b}{>{\hsize=2.0\hsize}X}

\begin{document}

% \title[Learning with Multi-Source Weak Social Supervision]{Learning with Multi-Source Weak Social Supervision \\\huge{- An Approach for Early Detection of Fake News}}% \\for Intent Detection}
\title{Leveraging Multi-Source Weak Social Supervision for Early Detection of Fake News}
\titlerunning{Early Detection of Fake News with Weak Social Supervsion}

% \title{{\m}: Learning with Multiple Sources of Supervision and User Interactions}

\author{Kai Shu\inst{1}, Guoqing Zheng\inst{2}, Yichuan Li\inst{1}, Subhabrata Mukherjee\inst{2}, Ahmed Hassan Awadallah\inst{2}, Scott Ruston\inst{1} \and Huan Liu\inst{1}}
\institute{
Computer Science and Engineering, Arizona State University, Tempe, AZ, USA
\email{\{kai.shu, yichuan1, scott.ruston, huan.liu\}}@asu.edu\and
Microsoft Research, Redmond, WA, USA\\
\email{\{guoqing.zheng, subhabrata.mukherjee, hassanam\}@microsoft.com}
}

\authorrunning{Anonymous}
% \keywords{Weak social supervision, fake news detection, machine learning}
%\keywords{Intent detection, weak supervision, communication understanding}

\maketitle
\begin{abstract}
Social media has greatly enabled people to participate in online activities 
at an unprecedented rate. However, this unrestricted access also exacerbates  
the spread of misinformation and fake news online which might cause confusion and chaos unless being detected early for its mitigation. Given the rapidly evolving nature of news events and the limited amount of annotated data,  state-of-the-art systems on fake news detection face challenges due to the lack of large numbers of annotated training instances that are hard to come by for early detection. %/  a supervised approach of feature engineering on news contents or embedding auxiliary information from social media for prediction, which may not be applicable to early detection of fake news. 
In this work, we %leverage weak supervision to address this challenge. Specifically, we 
exploit multiple weak signals from different sources given by user and content engagements (referred to as weak social supervision), and their complementary utilities to detect fake news. We jointly leverage limited amount of clean data along with weak signals from social engagements to train deep neural networks in a meta-learning framework to estimate the quality of different weak instances. %To enable early detection, we use only the news content as input for prediction without any social engagements. 
Experiments on real-world datasets demonstrate that the proposed framework outperforms  state-of-the-art  baselines  for early detection of fake news without  using  any  user engagements at prediction time.

\end{abstract}
\section{Introduction}

{\noindent \bf Motivation.} Social media platforms provide convenient means for users to create and share diverse information. Due to its massive content and convenience for access, more people seek out and receive news information online. For instance, around 68\% of U.S. adults consumed news from social media in 2018, a massive increase from corresponding 49\% consumption in 2012\footnote{https://bit.ly/39zPnMd} according to a survey by the Pew Research Center. However, social media also proliferates a plethora of misinformation and  fake news, i.e., news stories with intentionally  false information~\cite{shu2017fake}. Research has shown that fake news spreads farther, faster, deeper, and more widely than true news~\cite{vosoughi2018spread}. For example, during the 2016 U.S. election, the top twenty frequently-discussed false election stories generated 8.7 million shares, reactions, and comments on Facebook, more than the total of 7.4 million shares of top twenty most-discussed true stories\footnote{\url{https://bit.ly/39xmXT7}}.
% \begin{wrapfigure}{l}{0.65\textwidth}
%   \centering
%   {\includegraphics[width=0.60\textwidth]{example.png}
%   \caption{An illustration of a piece of fake news and related user engagements, which can be used for extracting weak social supervision. Users have different credibility, perceived bias, and express diverse sentiment to the news.}
%   \label{fig:illu}}
% \end{wrapfigure}
Widespread fake news can erode the public trust in government and professional journalism and lead to adverse real-life events. %For example, fake news claiming that Barack Obama was injured in an explosion wiped out \$130 billion in stock value in a short period of time\footnote{\url{https://www.forbes.com/sites/kenrapoza/2017/02/26/can-fake-news-impact-the-stock-market/##4986a6772fac}}. 
Thus, a timely detection of fake news on social media is critical to cultivate a healthy news ecosystem.  

{\noindent \bf Challenges.} %However, this is challenging for the following reasons. 
First, fake news is {diverse} in terms of topics, content, publishing method and media platforms, and sophisticated linguistic styles geared to emulate true news. Consequently, training machine learning models on such sophisticated content requires {\em large-scale annotated fake news data} that is egregiously difficult to obtain. Second, it is important to detect fake news {early}. Most of the research on fake news detection rely on signals that require a long time to aggregate, making them unsuitable for {\em early detection}. Third, the {evolving nature} of  fake news makes it essential to analyze it with signals from multiple sources to better understand the context. A system solely relying on social networks and user engagements can be easily influenced by biased user feedback, whereas relying only on the content misses the rich auxiliary information from the available sources. In this work, we adopt an approach designed to address the above challenges for early detection of fake news with limited annotated data by leveraging weak supervision from multiple sources involving users and their social engagements -- referred to as {\em weak social supervision} in this work.

%For newly emerging and time-critical events with less makes it non-trivial to exploit the rich auxiliary information signals directly. Fake news is related to newly emerging, time-critical events, which may not have been verified by existing knowledge bases or sites due to the lack of corroborating evidence. Moreover, detecting fake news at an early stage requires the prediction models to utilize minimal information from user engagements because extensive user engagements indicate more users are already affected by fake news.

\begin{figure}[tbp!]
    \centering
    {\includegraphics[width=0.65\textwidth]{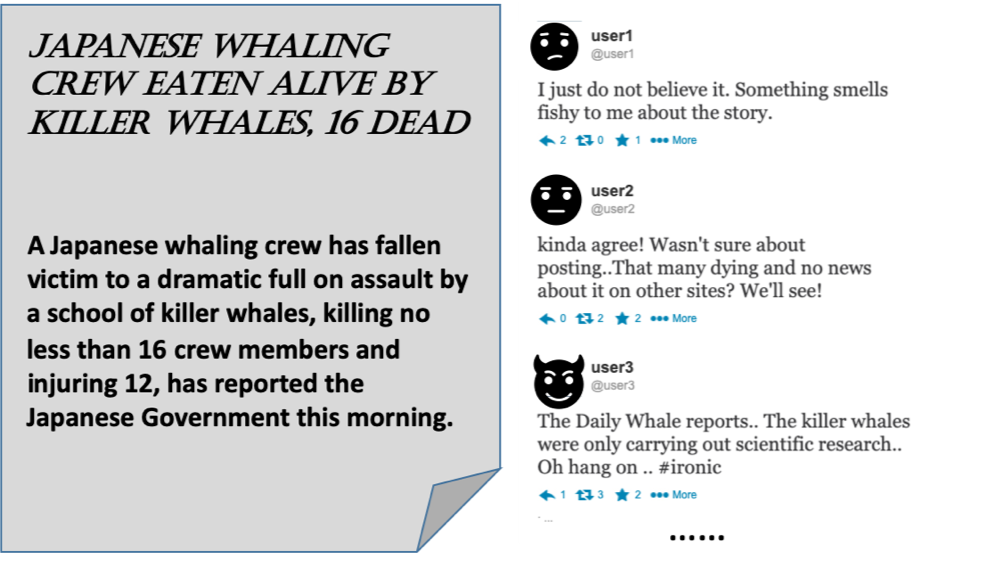}
    \caption{An illustration of a piece of fake news and related user engagements, which can be used for extracting weak social supervision. Users have different credibility, perceived bias, and express diverse sentiment to the news.}
    \label{fig:illu}}
    \vspace{-2em}
\end{figure}

%{\noindent \bf State-of-the-art and differences.} 
{\noindent \bf Existing work.} Prior works on detecting fake news~\cite{qian2018neural,wang2018eann} rely on large amounts of labeled instances to train supervised models. Such large labeled training data is difficult to obtain in the early phase of fake news detection. To overcome this challenge, learning with weak supervision presents a viable solution%been of great interest to the research community for various tasks
~\cite{zhang2019learning}. Weak signals are used as constraints to regularize prediction models~\cite{stewart2017label}, %learn with weak labels by re-weighting training instances~\cite{ren2018learning} 
or as %incorporate
loss correction mechanisms~\cite{hendrycks2018using}. Often only a \textit{single} source of weak labels is used. 

%However, social media data is multi-faceted with heterogeneous relationships between news content and propagators on social media. Many of the prior works on fake news have ignored this complex social link 
Existing research has focused either on the textual content relying solely on the linguistic styles in the form of sentiment, bias and psycho-linguistic features~\cite{pennebaker2015development} or on  tracing user engagements  on how fake news propagate through the network~\cite{wang2018eann}. 
In this work, we utilize {\em weak social supervision} to address the above shortcomings. Consider the example in Figure~\ref{fig:illu}. Though it is difficult to identify the veracity considering the news content in isolation, the surrounding context from other users' posts and comments provide clues, in the form of opinions, stances, and sentiment, useful to detect fake news. For example, in Figure~\ref{fig:illu}, the phrase ``\texttt{kinda agree..}'' indicates a positive sentiment to the news, whereas the phrase ``\texttt{I just do not believe it...}'' expresses a negative sentiment. Prior work has shown conflicting sentiments among propagators to indicate a higher probability of fake news~\cite{jin2016news,shu2017fake}. Also, users have different credibility degrees in social media and   
%Second, users in the social media have different credibility levels, where less-credible users %, such as malicious accounts or normal users who are vulnerable to fake news, 
less-credible ones are more likely to share fake news~\cite{shu2019beyond}. Although we do not have this information a priori, we can consider {\em agreement} between users as a weak proxy for their credibility. All of the aforementioned signals  from different sources like content and social engagements can be leveraged as weak supervision signals to train machine learning models. 

{\noindent \bf Contributions.} We leverage weak social supervision %in the form of weak signals from social context and textual content 
to detect fake news from limited annotated data. %Our models leverage the above signals for learning only during the training; whereas, at prediction time we use only the textual content --- thereby, allowing our models to detect fake news early without relying on the social context as features for prediction.
%These findings %heuristics and rules 
%from social media have great potential to bring additional signals to early detection of fake news. Thus, we propose to utilize and learn with multi-source of weak supervision simultaneously (in the form of weak labels) from social media to advance early fake news detection.
In particular, our model leverages a small amount of manually-annotated clean data and a large amount of weakly annotated data by proxy signals from multiple sources for joint training in a meta-learning framework. Since not all weak instances are equally informative, the models learn to estimate their respective contributions to optimize for the end task.  
%\todo{Subho: Add some details on the meta-learning framework}
%In this paper, we propose to leverage manually-annotated clean and multi-source noisy examples from social media {\em jointly} for early fake news detection. 
%Specifically, besides the fake news classifier that maps a news piece to the labels, 
To this end, we develop a Label Weighting Network (LWN) to model the weight of these weak labels that regulate the learning process of the fake news classifier. The LWN is serves as a meta-model to produce weights for the weak labels and can be trained by back-propagating the validation loss of a trained classifier on a separate set of clean data. The framework is uniquely suitable for early fake news detection, because it (1) leverages rich weak social supervision to boost model learning in a meta-learning fashion; and (2) only requires the news content during the prediction stage without relying on the social context as features for early prediction. Our contributions can be summarized as:
\begin{itemize}[leftmargin=*]
    \item {\bf Problem.} We study a novel problem of exploiting weak social supervision for early fake news detection;
    \item {\bf Method.} We provide a principled solution, dubbed {\m} to learn from \underline{M}ultiple-sources of \underline{W}eak \underline{S}ocial \underline{S}upervision ({\m}) from multi-faceted social media data. Our framework is powered by meta learning with a Label Weighting Network (LWN) to capture the relative contribution of different weak social supervision signals for training;
    \item {\bf Features.} We describe how to generate weak labels from social engagements of users that can be used to train our meta learning framework for early fake news detection along with quantitative quality assessment.  
    \item {\bf Experiments.} We conduct extensive experiments %on real-world datasets 
 to demonstrate the effectiveness of the proposed framework for early fake news detection over competitive baselines.
\end{itemize}

\section{Modeling Multi-source Weak Social Supervision}\label{sec:model}%\todo{Better name}

User engagements over news articles, including posting about, commenting on or recommending the news on social media, bear implicit judgments of the users about the news and could serve as weak sources of labels for fake news detection. For instance, prior research has shown that contrasting sentiment of users on a piece of news article, and similarly different levels of credibility or bias, can be indicators of the underlying news being fake. However, these signals are noisy and need to be appropriately weighted for training supervised models. Due to the noisy nature of such social media engagements, we term these signals as \textit{weak social supervision}. In the next section, we provide more details on how to extract these signals from social media engagements, while we focus on how to model them in the current section.

To give a brief overview, we define heuristic labeling functions (refer to Section~\ref{sec:weak}) on user engagements to harvest such signals to weakly label a large amount of data. The weakly labeled data is jointly combined with limited amount of manually annotated examples to build a fake news detection system that is better than training on either subset of the data. We emphasize that multiple weak labels can be generated for a single news article based on different labeling functions and we aim to jointly utilize both the clean examples as well as multiple sources of weak social supervision in this paper. 

In this section, we first formulate the problem statement, and then focus on developing algorithms for the joint optimization of manually annotated clean and multi-source weakly labeled instances in a unified framework. %We defer the detailed discussion of designing weak labeling functions to the next section.

\subsection{Problem Statement}
Let $\mathcal{D}=\{x_i, y_i\}_{i=1}^{n}$ denote a set of $n$ news articles with manually annotated clean labels, with $\mathcal{X}=\{x_i\}_{i=1}^{n}$ denoting the news pieces and $\mathcal{Y}=\{y_i\}_{i=1}^{n}\subset\{0,1\}^n$ the corresponding clean labels of whether the news is fake or not. 
% Each news article $x_i=\{w_1^i,\cdots,w_{m_i}^i\}$ contains a sequence of $m_i$ words. 
In addition, there is a large set of unlabeled examples. Usually the size of the clean labeled set $n$ is smaller than the unlabeled set due to labeling costs.  For the widely available unlabeled samples, we can generate weak labels by using different labeling functions  based on \textit{social engagements}. For a specific labeling function %{\yl{should ${\tilde{X}^{(k)}}$ be $\tilde{X}$ }}
$g^{(k)}: \mathcal{X}^{(k)}\rightarrow \tilde{\mathcal{Y}}^{(k)}$, where $\mathcal{X}^{(k)}=\{{x}_j^{(k)}\}_{j=1}^{N}$ denotes the set of $N$ unlabeled messages to which the labeling function $g^{(k)}$ is applied and $\tilde{\mathcal{Y}}^{(k)}=\{\Tilde{y}_j^{(k)}\}_{j=1}^N$ as the resulting set of weak labels. This weakly labeled data is denoted by $\Tilde{\mathcal{D}}^{(k)}=\{{x}_j^{(k)}, \Tilde{y}_j^{(k)}\}_{j=1}^{N}$ and often $n\ll N$.  We formally define our problem as:

\begin{center}
\fbox{\parbox[c]{.95\linewidth}{\textbf{Problem Statement:}
Given a limited amount of manually annotated news data $\mathcal{D}$ and $K$ sets of weakly labeled data $\{\Tilde{\mathcal{D}}^{(k)}\}_{k=1}^K$ derived from $K$ different weak labeling functions based on weak social signals, learn a fake news classifier 
$f: \mathcal{X}\rightarrow \mathcal{Y}$ which generalizes well onto unseen news pieces.
}}
\end{center}

\subsection{Meta Label Weighting with Multi-source Weak Social Supervision}\label{sec:base}
% \begin{table}[htbp!]
% \centering \caption{Notation Table.}
% % \small
% \begin{tabular}{ll}
% \toprule
% Notation & Meaning \\
% \midrule
% $\mathcal{D}$ & set of instances with clean labels \\
% %\midrule
% $\Tilde{\mathcal{D}}^{(k)}$& set of instances with weak labels from source $k$ \\
% %\midrule
% % $\mathcal{D}'$ & set of instances with weak label correction\\
% %\midrule
% % $enc_\theta(\cdot)$& shared encoder for learning data representations \\
%  $h_\theta(\cdot)$& shared encoder for learning data representations \\
% %  %\midrule
% %  $\mathbf{C}$& label corruption matrix \\
%  %\midrule
%  $f_c$ & function for predicting clean labels\\
%   %\midrule
%  $f_{w_k}$ & function for predicting weak labels from source $k$ \\
%   %\midrule
% %  $f'$ & function for correcting weak labels\\
% \bottomrule
% \end{tabular} \label{tab:symbol}
% % \vspace{-0.25cm}
% \end{table}

\begin{figure}[tbp!]
	\vspace{-0.25cm}
	\centering
	\hspace{-0.4cm}
% 	\subfigure[{An illustration of user engagements for extracting weak social supervision.}]{
% 	\includegraphics[width=0.45\textwidth]{example.jpg}\label{fig:illu}}
	\subfigure[{\m} Classifier]{
	\includegraphics[width=0.31\textwidth]{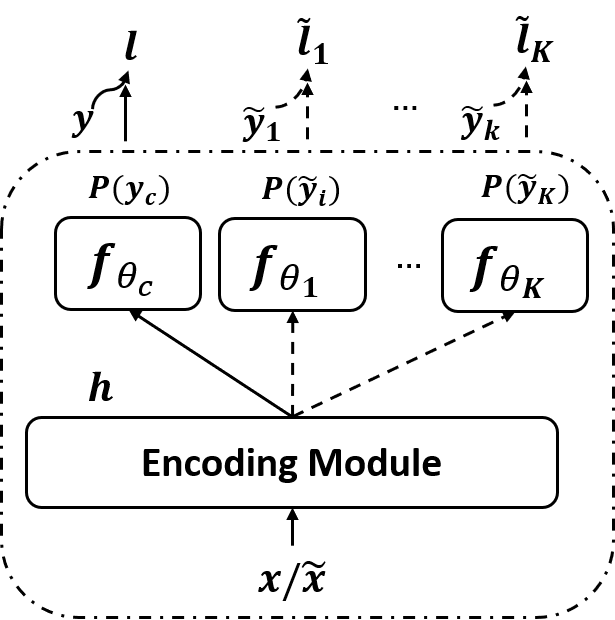}}
% 	\hspace{0.2cm}
    \subfigure[{\m} {\gn}]{
	\includegraphics[width=0.21\textwidth]{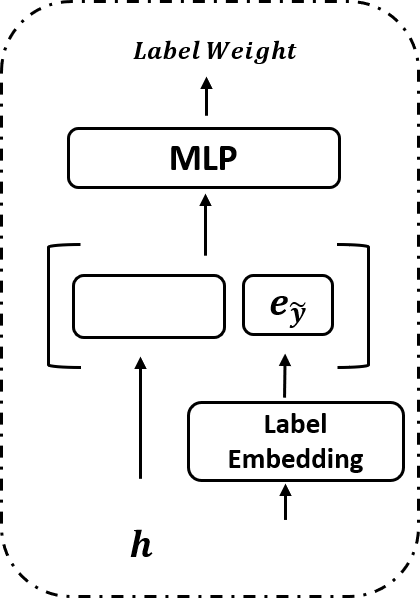}}
		\subfigure[{\m} inference]{
	\includegraphics[width=0.22\textwidth]{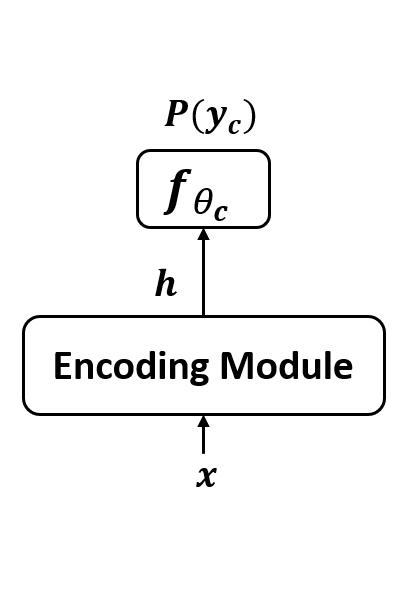}}
	\hspace{-0.6cm}
	\vspace{-0.2cm}
	\caption{The proposed framework {\m} for learning with multiple weak supervision from social media data. (a) Classifier: Jointly modeling clean labels and weak labels from multiple sources; (b) {\gn}: Learning the label weight based on the concatenation of instance representation and weak label embedding vector. (c) During inference, {\m} uses the learned encoding module and classification MLP $f_{w_{c}}$ to predict labels for (unseen) instances in the test data. %\kai{placeholder} 
	}\label{fig_framework}
	\vspace{-2em}
\end{figure}

Learning from multiple sources has shown promising performance in various domains such as truth discovery~\cite{ge2013multi}, object detection~\cite{ouyang2014multi}, etc. In this work, we have $K+1$ distinct sources of supervision: clean labels coming from manual annotation and multiple sources of weak labels obtained from $K$ heuristic labeling functions based on users' social engagements.

Our objective is to build an effective framework that leverages weak social supervision signals from multiple sources in addition to limited amount of clean data. However, signals from different weak sources are intrinsically noisy, biased in different ways, and thus of varying degree of qualities. Simply treating all sources of weak supervision as equally important and merging them to construct a single large set of weakly supervised instances tend to result in sub-optimal performance (as used as a baseline in our experiments). However, it is challenging to determine the contribution of different sources of weak social supervision. %{\yl{One more ablation study about considering the group source }}
%{\color{red} Defining the importance for each source as hyper-parameter and tuning them by cross-validation is one way, however this is ad-hoc in the first place and further it prohibits modeling more fine-grained importance weighting on instance level.}\todo{Subho: Remove this unless you have a corresponding baseline to show that this is indeed the case.}
To facilitate a principled solution of weighting weak instances, we leverage meta-learning. In this, we propose to treat label weighting as a meta-procedure, i.e.,  building a \textit{label weighting network} (LWN) which takes an instance (e.g., news piece) and its weak label (obtained from social supervision) as input, and outputs a scalar value as the importance weight for the pair. The weight determines the contribution of the weak instance in %optimizing for the end task, i.e., 
training the desired fake news classifier in our context. The LWN can be learned by back-propagating the loss of the trained classifier on a separate clean set of instances. To allow information sharing among different weak sources, for the fake news classifier, we use a shared feature extractor to learn a common representation and use separate functions (specifically, MLPs) to map the features to different weak label sources.

Specifically, let $h_{\theta_E}(x)$ be an encoder that generates the content representation of an instance $x$ with parameters $\theta_E$. Note that this encoder is shared by instances from both the clean and  multiple weakly labeled sources. Let $f_{\theta_c}(h(x))$ and $\{f_{\theta_k}(h(x))\}_{k=1,..,K}$ be the $K+1$ labeling functions that map the contextual representation of the instances to their labels on the clean and $K$ sets of weakly supervised data, respectively. In contrast to the encoder with shared parameters $\theta_E$, the parameters $\theta_c$ and $\{\theta_k\}_{k=1,...,K}$ are different for the clean and weak sources (learned by separate source-specific MLPs) to capture different mappings from the contextual representations to the labels from each source. 

For training, we want to jointly optimize the loss functions defined over the (i) clean data and (ii) instances from the weak sources weighted by their respective utilities. The weight of the weak label $\tilde y$ for an instance $x$ (encoded as $h(x)$) is determined by a separate Label Weighting Network (LWN) formulated as $\omega_\alpha(h(x),\Tilde y)$ with parameters $\alpha$. Thus, for a given $\omega_\alpha(h( x), \tilde y)$, the objective for training the predictive model with multiple sources of supervision jointly is:
\begin{align}
    \min_{\theta_E, \theta_c, \theta_1,...,\theta_k} &\mathbb{E}_{(x,y)\in\mathcal{D}}\ell(y,f_{\theta_c}(h_{\theta_E}(x)))%\nonumber\\
    +\sum_{k=1}^K\mathbb{E}_{({x},\Tilde{y})\in\mathcal{\Tilde{D}}^{(k)}}\omega_\alpha(h_{\theta_E}({x}),\Tilde{y})\ell(\Tilde{y},f_{\theta_k}(h_{\theta_E}({x})))
    \label{eq:L_train}
\end{align}
where $\ell$ denotes the loss function to minimize the prediction error of the model. The first component in the above equation optimizes the loss over the clean data, whereas the second component optimizes for the weighted loss (given by $w_\alpha(.)$) of the weak instances from $K$ sources. Figure \ref{fig_framework} shows the formulation for both the classifier and LWN.

\begin{algorithm}[tbp!]
 \caption{Training process of {\m}}
 \While{not converged}{
   1. Update LWN parameters $\alpha$ by descending $\nabla_{\alpha}\mathcal{L}_{val}(\theta-\eta\nabla_{\theta}\mathcal{L}_{train}( \alpha,  \theta))$\\
   2. Update classifier parameters $\theta$ by descending $\nabla_{\theta}\mathcal{L}_{train}(\alpha, \theta)$
 }
 \label{alg:train}
      \vspace{-0.2cm}
\end{algorithm}

 \begin{figure}[tbp!]
    \centering
    {
    \subfigure[{\gn} Update]{
	\includegraphics[width=0.65\textwidth]{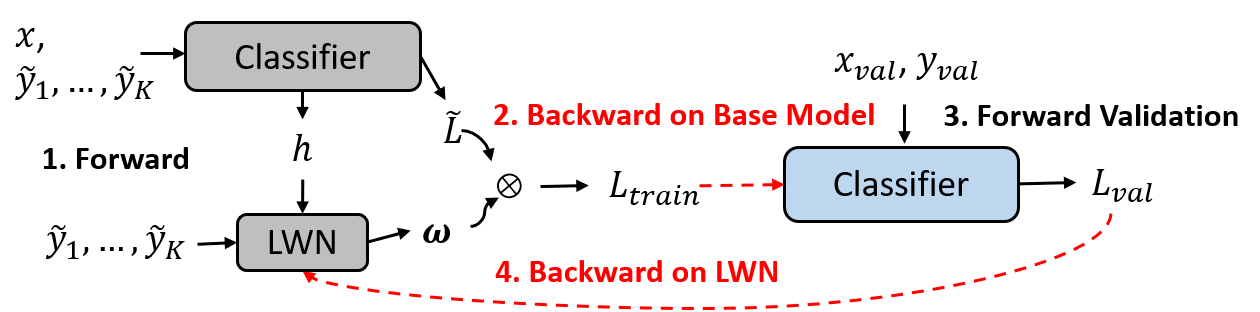}}
	\subfigure[Classifier Update]{
	\includegraphics[width=0.32\textwidth]{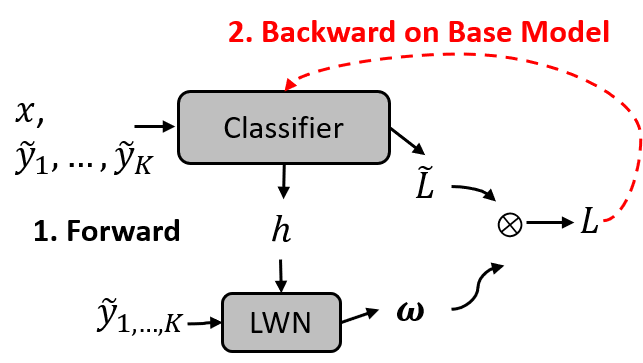}}
	
    \caption{The illustration of the {\m} in two phases: (a) we compute the validation loss based on the validation dataset and retain the computation graph for {\gn} backward propagation; 
    (b) the classifier update its parameters through backward propagation on clean and weakly labeled data. Note that $h$ is the set of hidden representations of the input instances, \textbf{$\mathbf{\omega}$} is the weight for each pair of instances and labels and $\otimes$ is point-wise multiplication. Gray indicates the parameters from the last iteration, blue indicates temporary updates. 
    % \kai{Gray indicates parameters from the last iteration, blue indicates temporary updates and yellow indicates parameters updated for the next iteration. }
     }
    \label{fig:illu1}}
    \vspace{-2em}
\end{figure}

The final objective is to optimize LWN $\omega_\alpha(h(x), \tilde y)$ such that when using such a weighting scheme to train the main classifier as specified by Eq. (\ref{eq:L_train}), the trained classifier could perform well on a separate set of clean examples. Formally, the following bi-level optimization problem describe the above intuition as:
\begin{align}
&\min_\alpha \mathcal{L}_{val}(\theta^*(\alpha))\,\,
\mbox{ s.t. }\,\,\theta^*=\arg\min\mathcal{L}_{train}(\alpha, \theta)
%&\min_\alpha \mathcal{L}_{val}(\theta_e^*(\alpha),\theta_c^*(\alpha))\\
%\mbox{ s.t. }\,\,&(\theta_e^*, \theta_c^*, \theta_1^*,...,\theta_K^*)=\arg\min\mathcal{L}_{train}(\alpha, \theta_e, \theta_c, \theta_1, ..., \theta_K)
\end{align}
where $\mathcal{L}_{train}$ is the objective in Eq. (\ref{eq:L_train}), $\theta$ denotes the concatenation of all classifier parameters $(\theta_E, \theta_c, \theta_1,...,\theta_K)$,  and $\mathcal{L}_{val}$ is loss of applying a trained model on a separate set of clean data. Note that $\theta^*(\alpha)$ denotes the dependency of $\theta^*$ on $\alpha$ after we train the classifier on a given LWN.
%\begin{align}
%\mathcal{L}_{val}=\mathbb{E}_{(x,y)}\ell(y,f_c(h(x)))
%\end{align}
    
%\todo{Subho: Things to clarify. (1) Val, train in L-val/L-train? (2) What is c*(a)? Note c is not a function, but a set of parameters. (3) Check the derivation and explain. Why is the first term zero in Eqn 4?}
%\todo{Guoqing: (1) I added the lines for L-train/L-val (2) C is a set of parameters, but after the model is trained, they depend on a, hence it's actually a function of a. (3) The derivation is correct based on derivatives of function compositions}

Analytically solving for the inner problem is typically infeasible. In this paper, we adopt the following one-step SGD update to approximate the optimal solution. As such the gradient for the meta-parameters $\alpha$ can be estimated as:
\begin{align}
     &\nabla_{ \alpha}\mathcal{L}_{val}( \theta-\eta\nabla_{\theta}\mathcal{L}_{train}(\alpha,\theta))  =-\eta \nabla_{ \alpha,   \theta}^2\mathcal{L}_{train}( \alpha, \theta)\nabla_{ \theta'}\mathcal{L}_{val}( \theta')
\\ \approx& -\frac{\eta}{2\epsilon}\left[\nabla_{ \alpha}\mathcal{L}_{train}( \alpha,  \theta^+)-\nabla_{ \alpha}\mathcal{L}_{train}( \alpha, \theta^-)\right]
\end{align}
where $ \theta^{\pm}= \theta\pm \epsilon\nabla_{ \theta'}\mathcal{L}_{val}( \theta')$, $ \theta'=\theta-\eta\nabla_{\theta}\mathcal{L}_{train}( \alpha, \theta)$, $\epsilon$ is a small constant for finite difference and $\eta$ is learning rate for SGD.

% We use back-propagation to train the neural networks. 
Since we leverage Multiple Weak Social Supervision, we term our method as {\m}. We adopt Adam with mini-batch training to learn the parameters. 
%Adadelta is an adaptive learning rate method which divides the learning rate by an exponentially decaying average, and is less sensitive to the initial learning rate. %We choose Adadelta as the optimizer because it is a popular and effective methods for determining the learning abortively, which is widely used for training neural networks.
% For ease of understanding, all the notations are summarized in Table~\ref{tab:symbol} and 
Algorithm \ref{alg:train} and Figure \ref{fig:illu1} outline the training procedure for {\m}. %\todo{Subho: This is the first reference for the system name. It should come earlier in Intro. First reference to Groupnet In Algorithm 1. Earlier referred as {\m}? Make system names consistent and introduce in Intro.}

\section{Constructing Weak Labels from Social Engagements}\label{sec:weak}
%\kai{better name for the title?}
In this section, we describe how to generate weak labels from users' social engagements that can be incorporated as weak sources in our model. We further perform evaluation to assess the quality of the weak labels.

\subsection{Dataset Description}\label{sec:data}
We utilize one of the most comprehensive fake news detection benchmark datasets called FakeNewsNet~\cite{shu2018fakenewsnet}. The dataset is collected from two fact-checking websites: GossipCop\footnote{https://www.gossipcop.com/} and PolitiFact\footnote{https://www.politifact.com/} containing news contents with labels annotated by professional journalists and experts, along with social context information. News content includes meta attributes of the news (e.g., body text), whereas social context includes related users' social engagements on the news items (e.g., user comments in Twitter). Note that the number of news pieces in PolitiFact data is relatively small, and we enhance the dataset to obtain more weak labels. Specially, we use a news corpus spanning the time frame 01 January 2010 through 10 June 2019, from 13 news sources including mainstream British news outlets, such as BBC and Sky News, and English language versions of Russian news outlets such as RT and Sputnik, which are mostly related to political topics. To obtain the corresponding social engagements, we use a similar strategy as FakeNewsNet~\cite{shu2018fakenewsnet} to get tweets/comments, user profiles and user history tweets through the  Twitter API
%~\footnote{https://developer.twitter.com/en/docs}
and web crawling tools. For GossipCop data, we mask part of the annotated data and treat them as unlabeled data for generating weak labels from the social engagements.
% \footnote{We will publicly release our processed datasets containing news content and social engagements as per the sharing policy of the platforms.} %Next, we introduce the details of weak labeling functions.

\subsection{Generating Weak Labels}
Now, we introduce the labeling functions for generating weak labels from social media via statistical measures guided by computational social theories. 
% Without loss of generosity, we introduce three rules for generating weak labels as follows, and leave other types of weak labeling rules for future work.

First, research shows user opinions towards fake news have more diverse sentiment polarity and less likely to be neutral~\cite{cui2019same}.  So we measure the sentiment scores (using a widely used tool VADER~\cite{hutto2014vader}) for all the users sharing a piece of news, and then measure the variance of the sentiment scores by computing the standard deviation. We define the following weak labeling function:
% \yl{For each r, the threhold is set by best classifying the clean data.}
% \subsubsection{Weak Labeling Functions from User Credibility}\label{sec:weak}
% \vspace{0.05in}
\begin{center}
\textit{\parbox[c]{.9\linewidth}{\textbf{Sentiment-based:}
% \yl{
% Standard deviation of users sentiments score. Sentiment score means users preference to the news. Standard deviation captures the variance and dispersion of users viewpoints towards this news. 
If a news piece has a standard deviation of user sentiment scores greater than a threshold $\tau_1$, then the news is weakly labeled as fake news. 
}}
\end{center}
% \vspace{0.05in}

Second, social studies have theorized the correlation between the bias of news publishers and the veracity of news pieces~\cite{gentzkow2015media}. 
% We can consider users who post news to be a proxy of the publishers with different degree of bias towards fake and real news~\cite{jin2016news}. 
Accordingly, we assume that news shared by users who are more biased are more likely be fake, and vice versa. Specifically, we adopt the method in~\cite{kulshrestha2017quantifying} to measure user bias (scores) by exploiting  users’ interests over her historical tweets. The hypothesis is that users who are more left-leaning or right-leaning share similar interests with each other. Following the method in~\cite{kulshrestha2017quantifying}, we generate representative sets of people with known public bias, and then calculate bias scores based on how closely a query users' interests match with those representative users.
% \todo{Subho: It is not clear how you compute this bias score. Requires more details.}  \kai{we revised accordingly}
% The bias score lies in range $[-1, 1]$ where $-1$ indicates left-leaning and $+1$ depicts right-leaning. 
We define the following weak labeling function:
% \vspace{0.05in}
\begin{center}
\textit{\parbox[c]{.9\linewidth}{\textbf{Bias-based:}
If the mean value of users' absolute bias scores -- sharing a piece of news -- is greater than a threshold $\tau_2$, then the news piece is weakly-labeled as fake news.}}
\end{center}
% \vspace{0.05in}

Third, studies have shown that less credible users, such as malicious accounts or normal users who are vulnerable to fake news, are more likely to spread fake news~\cite{shu2017fake}. %The credibility score depicts the trustworthiness of the user. 
To measure user credibility, we adopt the practical approach in~\cite{abbasi2013measuring}. The hypothesis is that less credible users are more likely to coordinate with each other and form big clusters, whereas more credible users are likely to form small clusters. We use the hierarchical clustering\footnote{\href{https://bit.ly/2WGK6zE}{https://bit.ly/2WGK6zE}} to cluster users based on their meta-information on social media and take the reciprocal of the cluster size as the credibility score. Accordingly, we define the following weak labeling function:
% The credibility scores are inferred from users' historical contents on social media. 

% \todo{Subho: It is not clear how you compute this credibility score. Requires more details.} \kai{we revised accordingly}
% \vspace{0.05in}
\begin{center}
\textit{\parbox[c]{.9\linewidth}{\textbf{Credibility-based:}
If a news piece has an average credibility score less than a threshold $\tau_3$, then the news is weakly-labeled as fake news.
}}
\end{center}
% \vspace{0.05in}

To determine the proper thresholds for $\tau_1$, $\tau_2$, and $\tau_3$, we vary the threshold values from $[0,1]$ through binary search, and compare the resultant weak labels with the true labels from the training set of annotated clean data -- later used to train our meta-learning model -- on GossipCop, and choose the value that achieves the the best accuracy on the training set. We set the thresholds as $\tau_1=0.15$, $\tau_2=0.5$, and $\tau_3=0.125$. 
Due to the sparsity for Politifact labels, for simplicity, we use the same rules derived from the GossipCop data. 

\noindent{\textbf{Quality of Weak Labeling Functions}}
% \subsection{}
We apply the aforementioned labeling functions and obtain the weakly labeled positive instances. We treat the news pieces discarded by the weak labeling functions as {\em negative} instances. The statistics are shown in Table~\ref{tab:data}. To assess the quality of these weakly-labeled instances, we compare the weak labels with the true labels on the annotated clean data in GossipCop -- later used to train our meta-learning model. %{\yl{Remove the confusion table. Both the accuracy and f1-score indicate the performance of the weak labeling functions }} Table~\ref{tab:confusion} shows the confusion matrix in Gossip Cop dataset with tuned thresholds as discussed before. 
The accuracy of the weak labeling functions corresponding to Sentiment, Bias, and Credibility are $0.59$, $0.74$, $0.74$, respectively. The F1-scores of these three weak labeling functions are $0.65$, $0.64$, $0.75$. We observe that the accuracy of the labeling functions are significantly better than random (0.5) for binary classification indicating that the weak labeling functions are of acceptable quality. 
% \todo{(1) The accuracy for using only Bias labeling function is greater than many of the baselines. What am I missing here? Is the recall the issue here with rule-based labeling functions? (2) We need to show the usefulness of using all labeling function together as opposed to only one of them for training the meta-learning model. I think the ablation test in Table 5 shows this individually, but not the performance on using all of them. Ideally, we should be better to show the value of multi-source supervision.}\kai{right, the results show that bias is a strong weak label; we add the results for all sources in table 5, which show better results}
%\todo{Subho: What are the f-scores for the weak labeling functions? Since these are rule-based, the recall (and hence the f-score) is likely to be lower than our model. If yes, then report them here}\kai{we added accordingly}

\begin{table}[tbp!]
\centering
\caption{The statistics of the datasets. Clean refers to manually annotated instances, whereas the weak ones are obtained by using the weak labeling functions% on a large set of unlabeled instances.
% \yl{preserve for gossipcop?}
}
\begin{tabular}{lccc}
\toprule
%   & \multicolumn{2}{c}{Enterprise Email} & \multicolumn{2}{c}{Digital Assistant}  \\
Dataset & Gossip Cop & Politifact\\
\midrule
  \# Clean positive & 1,546  & 303 \\
%\midrule
\# Clean negative & 1,546 & 303 \\

\# Sentiment-weak positive  & 1,894 & 3,067 \\
%\midrule
\# Sentiment-weak negative & 4,568  & 1,037 \\

%\midrule
\# Bias-weak positive  & 2,587 & 2,484 \\
%\midrule
\# Bias-weak negative & 3,875  & 1,620 \\
%\midrule
\# Credibility-weak positive  & 2,765 & 2,963 \\
%\midrule
\# Credibility-weak negative & 3,697  & 1,141 \\
% clean ratio & 0.111 & 0.111  \\
\bottomrule
\end{tabular} \label{tab:data}
\vspace{-0.4cm}
\end{table}

\section{Experiments}\label{sec:experiments}

In this section, we present the experiments to evaluate the effectiveness of {\m}. We aim to answer the following evaluation questions:
\begin{itemize}[leftmargin=*]
    \item \textbf{EQ1}: Can {\m} improve fake news classification performance by leveraging weak social supervision?
    \item \textbf{EQ2}: How effective are the different sources of supervision for improving prediction performance?
    \item \textbf{EQ3}: How robust is {\m} on leveraging  multiple sources?
\end{itemize}

% what is icew 
% extract social information about the tweets

\subsection{Experimental Settings}\label{sec:setting}

\noindent{\bf Evaluation measures.} We use F1 score and accuracy as the evaluation metrics. We randomly choose 15\% of the clean instances for validation %1,436 and 90 instances in GossipCop and PolitiFact dataset respectively, 
and 10\% for testing. We fix the number of weak training samples and select the amount of clean training data based on the {\em clean data ratio} defined as:
% \begin{equation*}
    $\text{clean ratio}=\frac{\text{\#clean labeled samples}}{\text{ \#clean labeled samples + \#weak labeled samples}}$.
% \end{equation*}
% 75\% of the instances in the clean set for training 718 and 456 instances.\todo{Subho: What are entities here? Requires clarification.} for GossipCop and PolitiFact respectively, 15\% for validation, and remaining 10\% for testing. \kai{we changed to instances }
This allows us to investigate the contribution of clean vs. weakly labeled data in later experiments. All the clean datasets are balanced with positive and negative instances. We report results on the test set with the model parameters picked with the best validation accuracy. All runs are repeated for 3 times and the average is reported. 

\noindent {\bf Base Encoders.} We use the convolutional neural networks (CNN)~\cite{kim2014convolutional} and RoBERTa-base, a robustly optimized BERT pre-training model~\cite{liu2019roberta} as the encoders for learning content representations. We truncate or pad the news text to 256 tokens, and for the CNN encoder we use pre-trained WordPiece embeddings from BERT to initialize the embedding layer. 
For each of the $K+1$ classification heads, we employ a two-layer MLP with 300 and 768 hidden units for both the  CNN  and RoBERTa encoders. The {\gn} contains a weak label embedding layer with dimension of 256, and a three-layer MLP with $(768, 768, 1)$ hidden units for each with a sigmoid as the final output function to produce a scalar weight between 0 and 1. We use binary cross-entropy as the loss function $\ell$ for {\m}\footnote{All the data and code will be available at this anonymous link.}.
%\footnote{All the data and code are available at: \href{https://www.dropbox.com/sh/3ku9p7emack4lmy/AAAbmrdcUr1yBrN5R-3c7-JAa?dl=0}{\textbf{this clickable anonymous link}}%\guoqing{new  URL?}
% }.

\noindent {\bf Baselines and learning configurations.} We consider the following settings: 

(1) training only with limited amount of manually annotated \textbf{clean} data. Models include the following state-of-the-art early fake news detection methods:
\begin{itemize}[leftmargin=*]
        \item {TCNN-URG}~\cite{qian2018neural}: This method exploits users' historical comments on news articles to learn to generate synthetic user comments. It uses a two-level CNN for prediction when user comments are not available for early detection.
    \item {EANN}~\cite{wang2018eann}: This method utilizes an adversarial learning framework with an event-invariant discriminator and fake news detector. For a fair comparison, we only use the text CNN encoder. 
\end{itemize}

(2) training only with \textbf{weakly} labeled data; and (3) training with both the \textbf{clean} and \textbf{weakly} labeled data as follows:
\begin{itemize}[leftmargin=*]
    \item {Clean+Weak}: In this setting, we simply merge both the clean and weak sets (essentially treating the weak labels to be as reliable as the clean ones) and use them together for training different encoders.
    \item {L2R}~\cite{ren2018learning}: L2R is the state-of-the-art algorithm for learning to re-weight (L2R) examples for training models through a meta learning process.
    \item {Snorkel}~\cite{alex2018training}: It combines multiple labeling functions given their dependency structure by solving a matrix completion-style problem. We use the label generated by Snokel as the weak label and feed it to the classification models. 
    % Snorkel uses a generative model for learning correlations across different weak labeling functions in the form of agreements and disagreements that are weighted to generate the latent true label.
    % \kai{add some description of how we use snorkel and train the label}
    \item {\m}: The proposed model for jointly learning with clean data and multi-sources of weak supervision for early fake news detection. 
    % \item \textbf{{\m}}: 
\end{itemize}
% For those methods involving weak labels, both single and multiple sources of weak labels can be applied. For the single source, it can be Sentiment, Bias, or Credibility.
Most of the above baseline models are geared for single sources. In order to extend them to multiple sources, we evaluated several aggregation approaches, and found that taking the majority label as the final label achieved the best performance result. We also evaluate an advanced multiple weak label aggregation method -- Snorkel~\cite{ratner2019snorkel} as the multi-source baseline. Note that our MWSS model, by design, aggregates information from multiple sources and does not require a separate aggregation function like the majority voting.
% For multiple sources, we consider two different aggregation approaches to consolidate the final prediction labels. The first approach performs a majority voting to get the final prediction results which is denoted as ``Majority-Vote". The second approach duplicates every instance for every weak functions and uses the bigger group as our final weak source data which is denoted as ``All-Source"
% \todo{Subho: This one is not clear. What do you mean by a bigger group?}. Initial experiments showed that ``Majority-Vote" works better than the ``All-Source", so we only keep the ``Majority-Vote" baseline without weight aggregation function, i.e weak, weak+clean, L2R. \todo{Subho: Remove All-source if you are not showing any results for this baseline. It does not add anything unless numbers are reported.}\kai{we changed accordingly}

% {\kai{\st{Adding the explanation is good. Please make sure you are consistently using the same term through our the experiments including figures and tables. You are using ``All Source'' in table 3, all vote in figure 3, etc.}}}
%
% When using weak rule S-B-C, for Weak and Clean+Weak, we first obtain the prediction results using the weak rule Sentiment, Bias, or Credibility, respectively; and then perform a majority voting to get the final prediction results.

\subsection{Effectiveness of Weak Supervision and Joint Learning}

\begin{table*}[t!]
\centering
\centering
	\caption{Performance comparison for early fake news classification. {\em Clean} and {\em Weak} depict model performance leveraging only those subsets of the data; {\em Clean+Weak} is the union of both the sets. }
	
% {\yl{\st{L2R and Snorkel are trained on both the clean and weak data {\yl{\st{with learned weights on instances from different sources}, Snorkel did not assign instance weight to each instance.}}. EANN and TCNN-URG are trained with only clean labels. Our method {\m} combining interactions from all sources outperforms all methods on all datasets.}}}}
		\label{tab:performance}
		\centering
		\begin{tabular}{@{}llccccccc@{}}
\toprule
%\rule{0pt}{10pt} 
\multirow{2}{*}{Methods} &  \multicolumn{2}{c}{GossipCop}  & &  \multicolumn{2}{c}{PolitiFact}\\ \cline{2-3} \cline{5-6} 

%\rule{0pt}{10pt} 
& \multicolumn{1}{c}{F1}  & \multicolumn{1}{c}{Accuracy} & & \multicolumn{1}{c}{F1} & \multicolumn{1}{c}{Accuracy}

\\ \midrule
TCNN-URG (Clean)  & 0.76 & 0.74 & & 0.77 & 0.78  \\
EANN (Clean) & 0.77 & 0.74 & & 0.78 & 0.81 \\
\midrule

CNN (Clean) & 0.74  & 0.73 & & 0.72 & 0.72 \\
CNN (Weak) & 0.73  & 0.65 & & 0.33 & 0.60 \\
CNN (Clean+Weak) & 0.76 & 0.74 & & 0.73 & 0.72 \\
CNN-\textit{Snorkel} (Clean+Weak) & 0.76 & 0.75 & & 0.78 & 0.73 \\
CNN-\textit{L2R} (Clean+Weak) & 0.77 & 0.74 & & 0.79 & 0.78 \\
%\midrule
CNN-{\m} (Clean+Weak) & \textbf{0.79 }& \textbf{0.77} && \textbf{0.82} & \textbf{0.82} \\
\midrule

RoBERTa (Clean)  & 0.77 & 0.76 & & 0.78 & 0.77 \\
RoBERTa (Weak)  & 0.74 & 0.74 & & 0.33 & 0.60 \\
RoBERTa (Clean+Weak)  & 0.80 & 0.79 & & 0.73 & 0.73\\
RoBERTa-\textit{Snorkel} (Clean+Weak)  & 0.76 & 0.74 & & 0.78 & 0.77 \\
RoBERTa-\textit{L2R} (Clean+Weak) & 0.78 & 0.75 & & 0.81 & 0.82 \\
%\midrule

RoBERTa-\textit{{\m}} (Clean+Weak)  & \textbf{0.80} & \textbf{0.80} & & \textbf{0.82} & \textbf{0.82} \\
\bottomrule
\end{tabular}

\end{table*}

To answer \textbf{EQ1}, we compare the proposed framework {\m} with the representative methods introduced in Section~\ref{sec:setting} for fake news classification. We determine the model hyper-parameters with cross-validation. For example, we set parameters $learning\_rate \in\{10^{-3}, 5\times 10^{-4},10^{-4},5\times 10^{-5}\}$ and choose the one that achieves the best performance on the held-out validation set. 
From Table~\ref{tab:performance}, we make the following observations:

\begin{itemize}[leftmargin=*]
    \item 
    Training only on clean data achieves better performance than training only on the weakly labeled data consistently  across all the datasets (clean $>$ weak).% hold for both base models including AvgEMb and BiLSTM, in RI, SM, and PA tasks.
    \item Among methods that only use clean data with CNN encoders, we observe TCNN-URG and EANN to achieve relatively better performance than Clean consistently. This is because TCNN-URG utilizes user comments during training to capture additional information, while EANN considers the event information in news contents (TCNN-URG$>$CNN-clean, and EANN$>$CNN-clean). 
    % \todo{Subho: Which Clean is this? CNN-Clean? Otherwise, observation not valid for Roberta-Clean.}\kai{yes, we revised}
    \item On incorporating weakly labeled data in addition to the annotated clean data, the classification performance improves compared to that using only the clean labels (or only the weak labels) on both datasets (demonstrated by clean+weak, L2R, Snorkel $>$ clean $>$ weak).
    \item On comparing two different encoder modules, we find that RoBERTa achieves much better performance in GossipCop compared to CNN, and has a similar performance in PolitiFact. The smaller size of the PolitiFact data %is much smaller than GossipCop (60 vs. 957 instances). This will cause 
    results in variable performance for RoBERTa.
    % {\yl{dataset inconsistency: In GossipCop, we use part of the data as weakly labeled data; In PoliticalFact, we use external data as weakly labeled data.}}
    
    \item For methods that leverage both the weak and clean data, L2R and Snorkel perform quite well. This is because L2R assigns weight to instances based on their contribution with a held-out validation set, whereas Snorkel leverages correlations across multi-source weak labeling functions to recover the label. 
    % \ms{Confusing way to refer to these techniques and inconsistent with how we have referred to them earlier.}
    % \item The performance of \m-Base using both clean and weak labels is better than using a subset of them (demonstrated by \m-Base $>$ Clean, \m-Base $>$ Weak). 
    \item In general, our model {\m} achieves the best performance. We observe that {\m} $>$ L2R and Snorkel on both the datasets. This demonstrates the importance of treating weak labels differently from the clean labels with a joint encoder for learning shared representation, separate MLPs for learning source-specific mapping functions, and learning to re-weight instances via {\gn}. To understand the contribution of the above model components, we perform an ablation study in the following section.
    % In addition, the performance of using all the weak labels (S-B-C) achieves better performance than using a single source, indicating the complementary information among the weak labels.
\end{itemize}

% \kai{may discuss the different setting with SIGIR paper here briefly}
% \vspace{-0.2cm}
\subsection{Impact of the Ratio of Clean to Weakly Labeled Data on Classification Performance}
\begin{figure*}[tbp!]
    \centering
    \subfigure[GossipCop]{
    \includegraphics[width=0.4\textwidth]{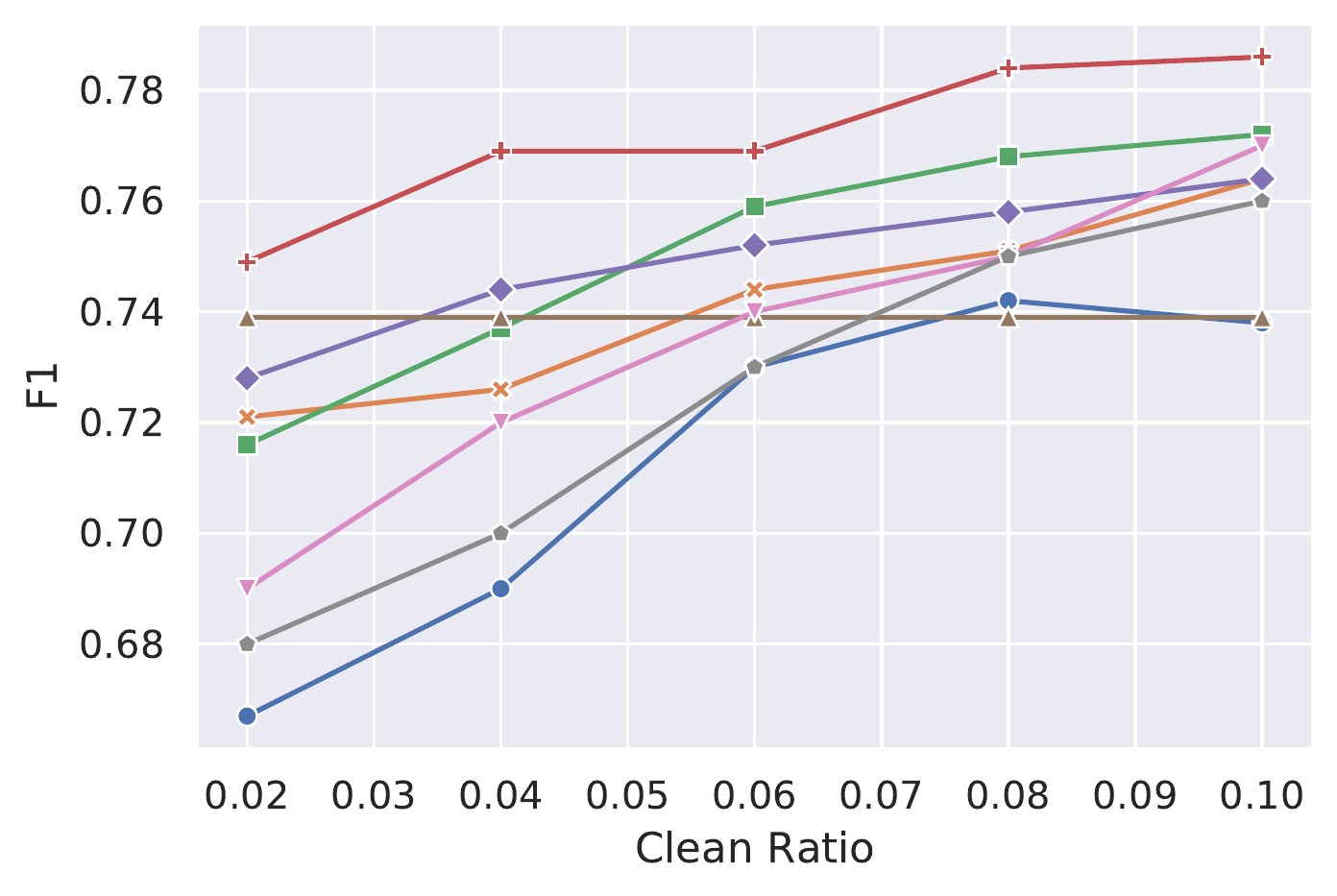}
    }
    \subfigure[PolitiFact]{
    \includegraphics[width=0.4\textwidth]{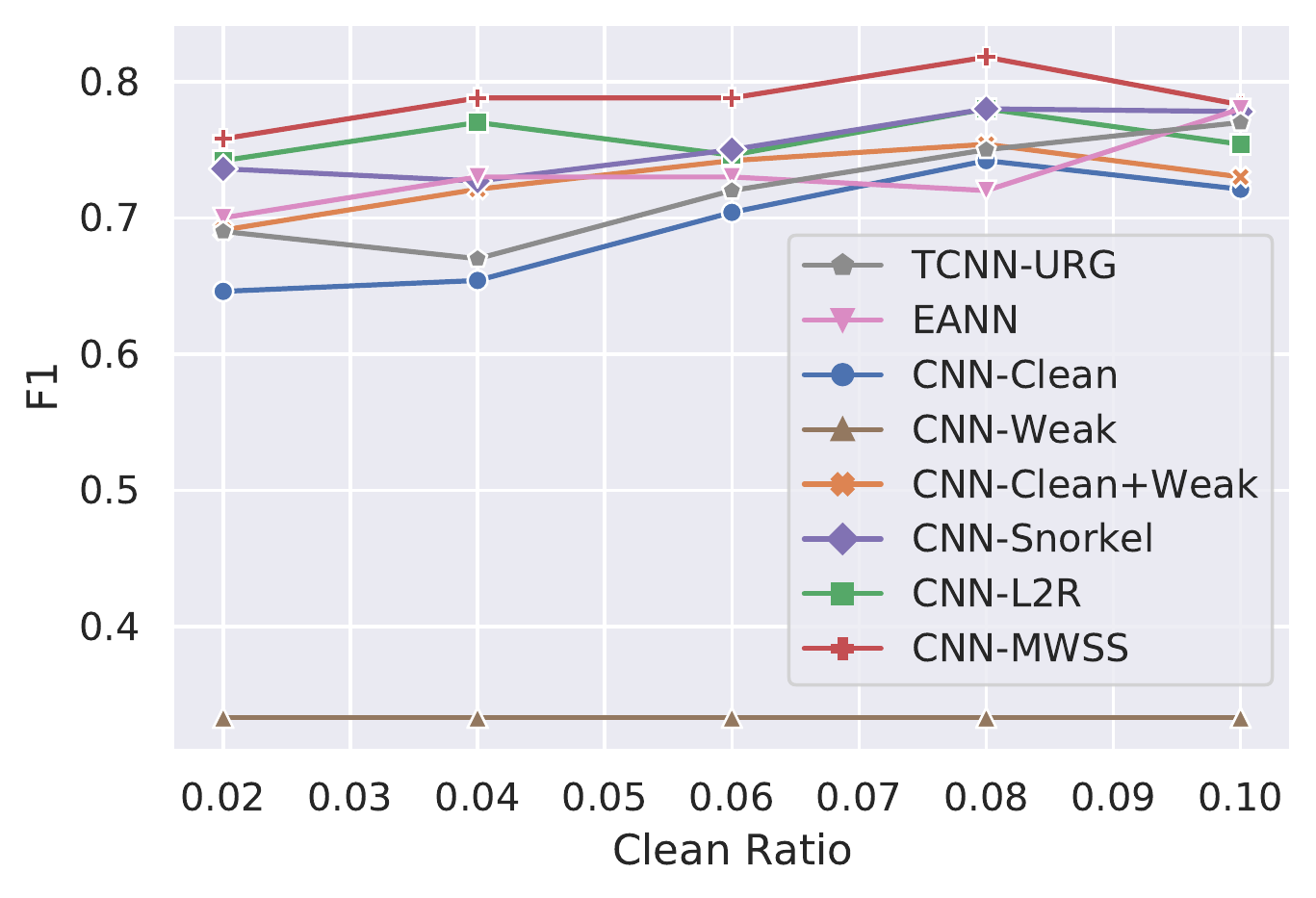}
    }
    \vspace{-1em}
    \caption{F1 score with varying clean data ratio from 0.02 to 0.1 with CNN-{\m}. The trend is the similar with RoBERTa encoder (best visualized in color).}
    \label{fig:golden_ratio}
    \vspace{-2em}
\end{figure*}

% In the real world scenario, we often have a limited amount of annotated data and a large amount of unlabeled data. 
To answer \textbf{EQ2}, we explore how the performance of {\m} changes with the clean ratio. We set the clean ratio to vary in $\{0.02,0.04,0.06,0.08, 0.1\}$. To have a consistent setting, we fix the number of weakly labeled instances and change the number of clean labeled instances accordingly. In practise, we have abundant weak labels from the heuristic weak labeling functions. The objective here is to figure out how much clean labels to add in order to boost the overall model performance. Figure~\ref{fig:golden_ratio} shows the results. We make the following observations: 
\begin{itemize}[leftmargin=*]
\item With increasing values of clean ratio, the performance increases for all methods (except \textit{Weak} which uses a fixed amount of weakly labeled data). This shows that increasing amount of reliable clean labels obviously helps the models.
\item For different clean ratio configurations, {\m} achieves the best performance compared to other baselines, i.e., {\m} $>$ L2R and Snorkel. This shows that {\m} can more effectively utilize the clean and weak labels via its multi-source learning and re-weighting framework.
\item We observe that the methods using Clean+Weak labels where we treat the weak labels to be as reliable as clean ones may not necessarily perform better than using only clean labels.
% \ms{Clean+Weak vs. Clean and weak could be confusing for the reader.} 
This shows that simply merging the clean and weak sources of supervision without accounting for their reliability may not improve the prediction performance. 
% \item The performance gap between using only the clean labels and {\m} using both clean and weak labels decreases with the increase in the clean ratio. Note that {\m} is more effective when the clean ratio is small which is a more realistic setting with small amount of labeled data.
\end{itemize}

\subsection{Parameter Analysis}
% In this section, we perform ablation studies to analyze the impact of multiple sources of weak labeling functions and different model components.
%respectively. 
\begin{table} [tbp!]
\centering 
% \vspace{-0.2cm}
\caption{F1/Accuracy on training {\m} on different weak sources with clean data.}
\begin{tabular}{ccccc}
\toprule
Dataset & Sentiment & Bias & Credibility & All Sources\\
\midrule
GossipCop  & 0.75/0.69 & 0.78/0.75 & 0.77/0.73 & 0.79/0.77 \\
\midrule
PolitiFact & 0.75/0.75 & 0.77/0.77 & 0.75/0.73 & 0.78/0.75 \\
\bottomrule%\hline
\end{tabular}  \label{tab:ab_source}
\vspace{-0.4cm}
\end{table}

\noindent\textbf{Impact of source-specific mapping functions:} %\todo{Subho: What do you mean by a head? I thought it is a transformer-head. Need a different terminology. Do you mean multiple MLPs for different sources?}
% -------------- Seperation
In this experiment, we want to study the impact of modeling separate MLPs for source-specific mapping functions (modeled by $f_{\theta_k}$ in Equation~\ref{eq:L_train}) in {\gn} and L2R as opposed to replacing them with a single shared MLP (i.e. $f_{\theta_k} = f_\theta \ \forall k$) across multiple sources.
From Table~\ref{tab:ab_head_weight}, we observe that {\m} and L2R both work better with multiple source-specific MLPs as opposed to a single shared MLP %than single head group net model and multiple head L2R works better than single head L2R. 
by better capturing source-specific mapping functions from instances to corresponding weak labels.
%This give us confidence thacorret multiple head can resolve the noise from the weak source data by passing different classifiers. 
We also observe {\m} to perform better than L2R for the respective MLP configurations -- demonstrating 
%The second observation is that the comparison between single head L2R and single head {\m}, and the comparison with multiple head L2R and {\m} prove 
the effectiveness of our re-weighting module. %From the reported result, we can observe that multiple classification head and the {\gn} module have superior performance than others. 

\noindent\textbf{Impact of different weak sources}:
\begin{table}[tp!]
% \vspace{-0.2cm}
	\begin{minipage}{0.58\linewidth}
	\caption{F1/Accuracy result of ablation study on modeling source-specific MLPs with different clean ratio (C-Ratio). ``SH" denotes a single shared MLP and ``MH" denotes multiple source-specific ones. } \label{tab:ab_head_weight}
	\centering
	\begin{tabular}{llcccc}
\toprule%hline
%TODO
Model & C-Ratio & L2R & {\gn}  \\
\midrule
\multirow{2}{*}{SH} & 0.02 & 0.72/0.68  &  0.73/0.72  \\
& 0.10 & 0.77/0.74 & 0.77/0.73 \\
\midrule
\multirow{2}{*}{MH} & 0.02 & 0.73/0.71 &  0.75/0.71  \\
& 0.10 & 0.78/0.76 & 0.79/0.77  \\
 \bottomrule
\end{tabular} 
\end{minipage}
\makeatletter\def\@captype{figure}\makeatother
	\begin{minipage}{0.42\linewidth}
	\vspace{0.5cm}
    \includegraphics[width=1\textwidth]{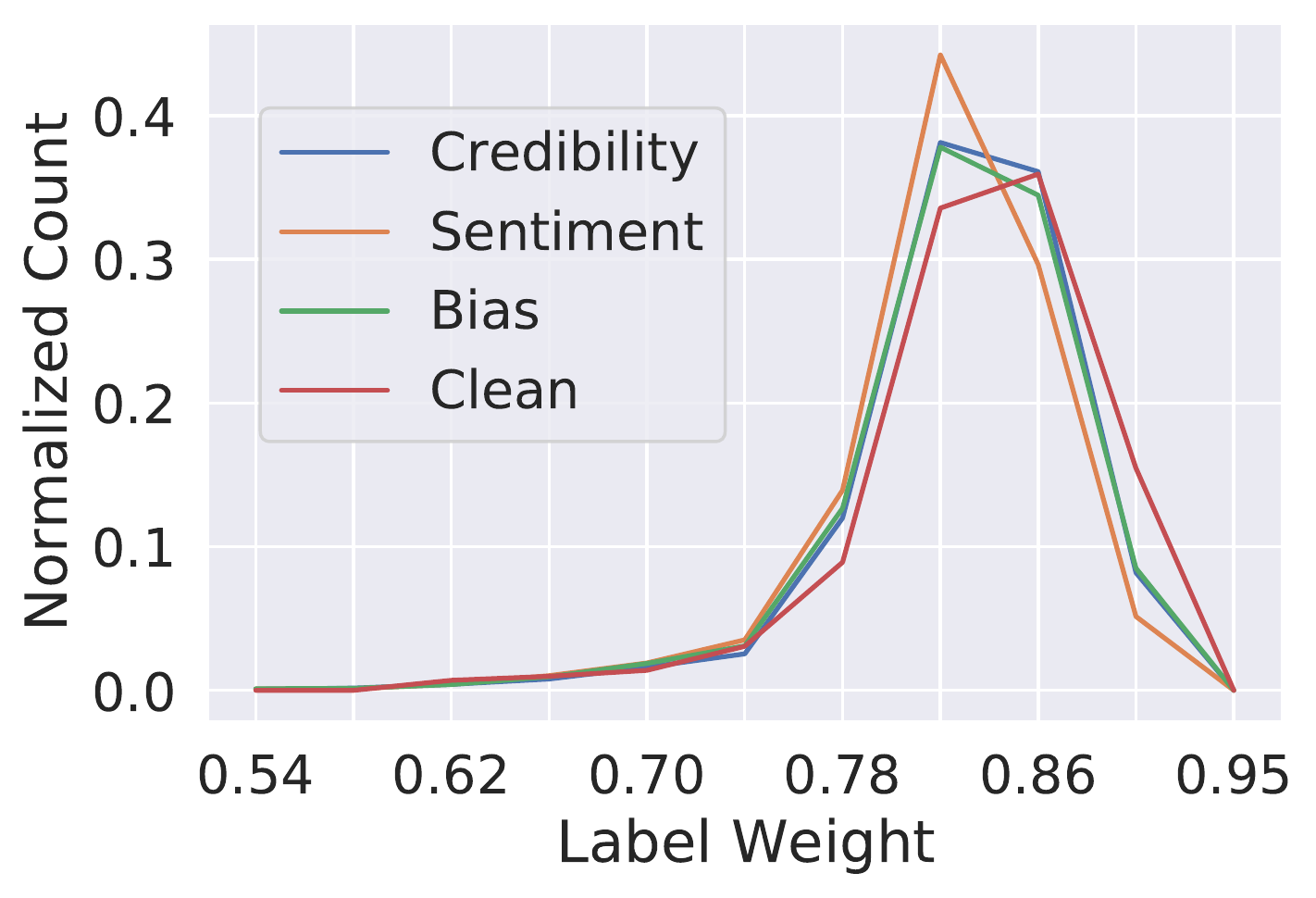}
		\caption{Label weight density distribution among weak and clean instances in GossipCop. The mean of the label weight for \textit{Credibility}, \textit{Sentiment}, \textit{Bias} and \textit{Clean} are 0.86, 0.85, 0.86 and 0.87 respectively.}
% 		The PolitiFact dataset has a similar trend.}
		\label{fig:ab_source}
	\end{minipage}
\hfill
\vspace{-3em}
\end{table}
To study the impact of multi-source supervision, we train {\m} separately with individual weak sources of data along with clean annotated instances with a clean ratio of $0.1$. From Table~\ref{tab:ab_source}, we observe that training {\m} with multiple weak sources achieves better performance compared to that of a single weak source -- indicating complementary information from different weak sources help the model. To test whether {\m} can capture the quality of each source, we visualize the label weight distribution for each weak source and clean dataset in Figure~\ref{fig:ab_source}. 
% We observe that sentiment viewpoints in these datasets mainly capture political ideologies (e.g., democratic vs. republican) and is not diverse enough to capture other forms of demographic bias.
 %\todo{Subho: Figure missing.}.
From the weight distribution, we also observe the weight of the sentiment-source (referred as {\em Sentiment}) to %gather in the left and the mean of weight is 
be less %\todo{Subho: How do you compute the weight of a source? Mean of weight of all instances from the source, or separate hyper-parameter? Requires clarification.} 
than that of other sources. In addition, although the {\gn} is not directly trained on clean samples, it still assigns the largest weight to the clean source. These demonstrate that our model not only learns the importance of different instances but also learns the importance of the corresponding source.

\section{Related Work} \label{sec:related} %\todo{Put this section in the end}
% In this section, we briefly review the related work on 1) Fake News Detection and 2) Learning with Weak Supervision.

% \subsection{Fake News Detection}
\noindent{\textbf{Fake News Detection:}}
Fake news detection methods mainly focus on using \textit{news contents} and with information from \textit{social engagements}~\cite{shu2017fake}. Content-based approaches exploit feature engineering or latent features such as using deep neural networks to capture deception cues~\cite{potthast2017stylometric,rubin2015truth}.
% For news content based approaches, features are extracted as using linguistic factors and visual factors. Linguistic features capture specific writing styles and sensational headlines that commonly occur in fake news content~\cite{potthast2017stylometric}, such as lexical and syntactic features. Visual-based features try to identify fake images~\cite{gupta2013faking} that are created or capture specific characteristics for images in fake news. News content-based models include (1) knowledge-based: using external sources to fact-checking claims in news content~\cite{magdy2010web,wu2014toward}, and (2) style-based: capturing the manipulators in deceptive or non-objectivity writing style~\cite{feng2012syntactic,rubin2015truth,potthast2017stylometric}.
For social context based approaches, features are mainly extracted from users, posts and networks. User-based features are extracted from user profiles to measure their characteristics~\cite{castillo2011information}. Post-based features represent users' response in term of stance, topics, or credibility~\cite{castillo2011information,jin2016news,ma2015detect}. Network-based features are extracted by constructing specific networks. Recently, deep learning models are applied to learn the temporal and linguistic representation of news~\cite{qian2018neural,wang2018eann}.  Wang~\textit{et al.}
proposed an adversarial learning framework with an event-invariant discriminator and fake news detector~\cite{wang2018eann}.
% Guo~\textit{et al.} proposed to use social features to guide the hierarchical comments networks to learn representations for fake news detection~\cite{guo2018rumor}. 
Qian~\textit{et al.} proposed to use convolutional neural networks to learn representations from news content, and a conditional variational auto-encoder to capture features from user comments for training, and only use news contents with generated comments for early fake news detection~\cite{qian2018neural}.

% \subsection{Learning with Weak Supervision}
% \kai{Rephrase the text, add L2R, label correction}
\noindent{\textbf{Learning with Weak Supervision:}}
Most machine learning models rely on the quality of labeled data to achieve good performance where the presence of label noise or adversarial noise cause a dramatic performance drop~\cite{reed2014training}. Therefore, learning with noisy labels has been of great interest for various tasks~\cite{natarajan2013learning,zhang2019learning}. Existing works attempt to rectify the weak labels by incorporating a loss correction mechanism~\cite{sukhbaatar2014training,patrini2017making,li2017learning}. Sukhbaatar \textit{et al.}~\cite{sukhbaatar2014training} introduce a linear layer to adjust the loss and estimate label corruption with access to the true labels~\cite{sukhbaatar2014training}. Patrini \textit{et al.}~\cite{patrini2017making} utilize the loss correction mechanism to estimate a label corruption matrix without making use of clean labels. Other works consider the scenario where a small set of clean labels are available~\cite{hendrycks2018using,ren2018learning,zheng2019meta}. For example, Zheng \textit{et al.} propose a meta label correction approach using a meta model which provides reliable labels for the main models to learn.
% li2017learning,charikar2017learning,
% Veit \textit{et al.} use human-verified labels and train a label cleaning network in a multi-label classification setting. 
Recent works also consider that weak signals are available from multiple sources~\cite{ratner2017snorkel,varma2019learning,ratner2018training} to exploit the redundancy as well as the consistency in the labeling information. 
% In contrast, the weak signals in our work are derived from rich user interactions. In addition, our work does not make any strong assumptions about the structure of the noise or depend on the availability of multiple weak sources to model corroboration. 
% User interaction has been widely used to build and improve Web search and recommender systems, including using user clicks to improve data quality for learning to rank~\cite{xu2010improving}, predicting and improving search results~\cite{agichtein2006learning,agichtein2006improving}, and incorporating user interactions for better recommendation ~\cite{rendle2010pairwise,jiang2012social}.
In contrast to all these models trained on manually annotated clean labels, we develop a framework {\m} that leverages weak social supervision signals from rich social media for early fake news detection in addition to a small amount of clean labels.

% \kai{merge 5.1 to 5.2 }

\section{Conclusions and Future Work}\label{sec:conclusion}
In this paper, we develop techniques for early fake news detection leveraging weak social supervision signals from multiple sources. We develop an end-to-end neural network model {\m} to jointly learn from limited amount of clean annotated labels and large amount of weakly labeled instances from different sources. Our framework is powered by meta learning with a Label Weighting Network (LWN) to capture the varying importance weights of such weak supervision signals from multiple sources during training.
% We demonstrate the effectiveness of {\m} on early fake news detection with weak labels derived from social media interactions. 
Extensive experiments in real-world datasets show {\m} to outperform state-of-the-art baselines without using any user engagements at prediction time. 

As future work, we want to explore other techniques like label correction methods to obtain high quality weak labels to further improve our models. In addition, we can extend our framework to consider other types of weak social supervision signals from social networks leveraging temporal footprints of the claims and engagements.

%and, temporal information for detecting fake news and other tasks as well.

\bibliographystyle{plain}
\bibliography{ref,ref_fake}

\end{document}